\documentclass[lettersize,journal]{IEEEtran}
\usepackage{amsmath,amsfonts}
\usepackage{algorithmic}
\usepackage{algorithm}
\usepackage{array}
\usepackage[caption=false,font=normalsize,labelfont=sf,textfont=sf]{subfig}
\usepackage[colorlinks=true, linkcolor=blue, citecolor=blue, urlcolor=blue]{hyperref}
\usepackage{textcomp}
\usepackage{stfloats}
\usepackage{url}
\usepackage{verbatim}
\usepackage{graphicx}
\usepackage{cleveref}
\usepackage{cite}
\usepackage{xcolor,color}
\usepackage{booktabs}   
\usepackage{diagbox}    
\usepackage{multirow}   
\usepackage{makecell}
\usepackage{listings}
\usepackage[numbers]{natbib} 
\usepackage{tabularx}
\usepackage{bbding}
\usepackage{pifont}
\usepackage{changepage}
\usepackage{caption}
\setcitestyle{square,citesep={,}}
\newcommand{\jq}[1]{\textcolor[rgb]{0,0.0,0}{#1}}
\begin{document}

\title{Modality-Agnostic Prompt Learning for Multi-Modal Camouflaged Object Detection}

\author{Hao Wang, Jiqing Zhang, Xin Yang, Baocai Yin, Lu Jiang, Zetian Mi, Huibing Wang
\thanks{Hao Wang, Jiqing Zhang, Lu Jiang, Zetian Mi, and Huibing Wang are with the Information Science and Technology College, Dalian Maritime University, Liaoning, 116026, China (email: King.Whao@outlook.com; jqz@dlmu.edu.cn; jiangl761@dlmu.edu.cn; mizetian@dlmu.edu.cn;  huibing.wang@dlmu.edu.cn).}
\thanks{Baocai Yin is with the Beijing University of Technology, Beijing, 100124, China (email: ybc@dlut.edu.cn). }
\thanks{Xin Yang is with the Dalian University of Technology, Liaoning, 116024, China (email: xinyang@dlut.edu.cn).}
}

\markboth{Journal of \LaTeX\ Class Files,~Vol.~14, No.~8, August~2025}%
{Shell \MakeLowercase{\textit{et al.}}: A Sample Article Using IEEEtran.cls for IEEE Journals}


\maketitle

\begin{abstract}

Camouflaged Object Detection (COD) aims to segment objects that blend seamlessly into complex backgrounds, \jq{with growing interest in exploiting additional visual modalities to enhance robustness through complementary information. However, most existing approaches generally rely on modality-specific architectures or customized fusion strategies, which limit scalability and cross-modal generalization. To address this, we propose a novel framework that generates modality-agnostic multi-modal prompts for the Segment Anything Model (SAM), enabling parameter-efficient adaptation to arbitrary auxiliary modalities and significantly improving overall performance on the COD tasks. Specifically, we model multi-modal learning through interactions between a data-driven content domain and a knowledge-driven prompt domain, distilling task-relevant cues into unified prompts for SAM decoding. We further introduce a lightweight Mask Refine Module to calibrate coarse predictions by incorporating fine-grained prompt cues, leading to more accurate camouflaged object boundaries. Extensive experiments on RGB–Depth, RGB–Thermal, and RGB–Polarization benchmarks validate the effectiveness and generalization of our modality-agnostic framework.}

\end{abstract}


\section{Introduction}

\jq{Camouflaged Object Detection (COD) aims to segment objects that are highly blended into complex backgrounds \cite{fan2020camouflaged}, with broad applications in medical image analysis, industrial defect inspection, and remote sensing. Despite recent progress, RGB-only COD methods remain fragile under challenging camouflage conditions, where strong foreground-background similarity and weak boundary contrast hinder reliable foreground cue extraction. To improve robustness, multi-modal COD has gained increasing attention by leveraging complementary cues from auxiliary visual modalities such as depth \cite{wang2024ipnet}, thermal \cite{11159525}, and polarization \cite{wang2024ipnet} to enhance foreground-background separability.}

\def\w{0.8\linewidth}
\def\h{1.0in}
\begin{figure}[tbp]
    \setlength{\tabcolsep}{1.0pt}
    \centering
    \small
    \begin{tabular}{c}
        \includegraphics[width=\w]{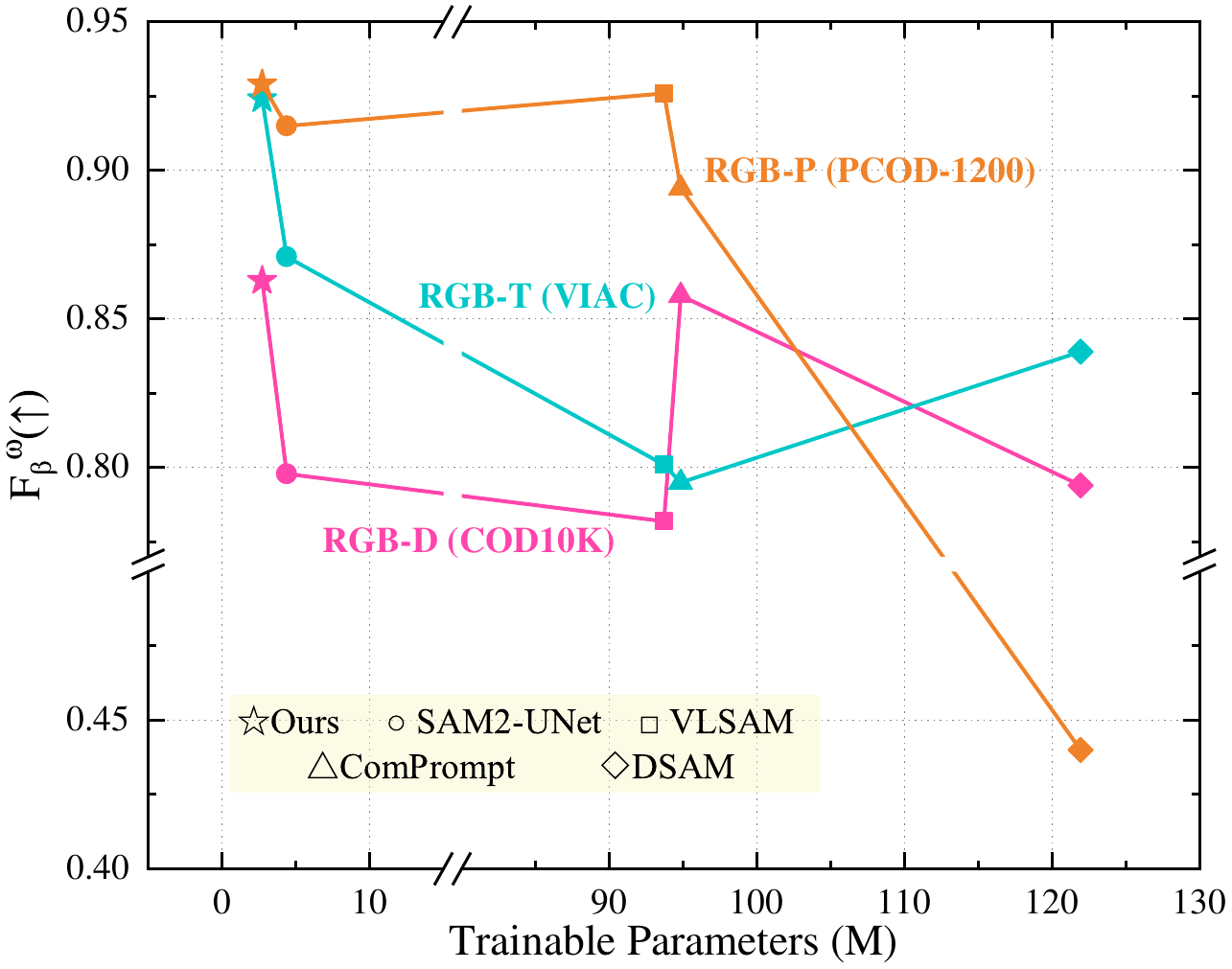}
    \end{tabular}
    \caption{The $F_{\beta}^{w}$ scores with respect to the trainable parameters
of different methods on  COD10K (RGB-Depth), PCOD-1200 (RGB-Polarization), and VIAC (RGB-Thermal) datasets.}
    \label{fig:Fbubble_params}
\end{figure}

\jq{Early multi-modal COD methods typically rely on carefully designed fusion strategies to exploit cross-modal complementarity, such as attention mechanisms \cite{wang2024ipnet} and dual-stream architectures \cite{wang2023depth}. Recently, driven by the strong generalization of the Segment Anything Model (SAM) as a large-scale pretrained visual foundation model \cite{kirillov2023segment}, an increasing number of studies have incorporated SAM into COD. Existing approaches typically adopt parameter-efficient adaptation, including fine-tuning a small subset of parameters \cite{zhang2025comprompter} or inserting lightweight adapters \cite{chen2023sam} to frozen SAM. While effective with low training cost, these approaches remain modality-specific, requiring dedicated architectural designs for each auxiliary modality, which limits scalability and cross-modal generalization. For example,  DSAM \cite{yu2024exploring} constructs a multi-stage framework with dual-branch embeddings, knowledge distillation, and bias correction to extract geometry-aware cues from depth and convert them into prompt representations. SAM-DSA \cite{liu2025improving} introduces a depth adapter and employs a two-stage knowledge distillation scheme to overcome inter-modal differences and suppress potential noise from depth images. Therefore, a natural question arises: \textit{can we design a more efficient and modality-agnostic approach that converts arbitrary auxiliary modalities into SAM-compatible prompt representations for scalable RGB+X camouflaged object segmentation?}
}

\jq{To this end, this paper proposes a novel modality-agnostic prompting framework that encodes auxiliary modalities into unified modality prompts and injects them into SAM decoding in a lightweight manner. We attribute our modality-agnostic property to a more general modeling paradigm: instead of designing modality-specific architectures or customized multi-modal fusion schemes, we focus on interactions between the content domain and the prompt domain. Specifically, the content domain is established by element-wise summation of RGB and auxiliary-modal features, providing heterogeneous target-related spatial cues and semantic information.
The prompt domain denotes prompt representations formed by input-adaptively modulating learnable prompt prototypes, encoding selective guidance signals for detection. The two domains correspond to data-driven candidate evidence and knowledge-driven priors, respectively. 
We therefore employ dual-path cross-attention to establish a bidirectional synergy between the content and prompt domains. This mechanism distills task-relevant evidence from heterogeneous observations into prompt prototypes while rectifying spatial fidelity via structured priors. To enhance representational flexibility, we also introduce a set of learnable Auxi-tokens that are concatenated with SAM’s original output and prompt tokens. Moreover, a Mask Refine module is introduced for predictive calibration. By synergizing the coarse predictions from the Mask Decoder with fine-grained prompts, this module further enhances the overall precision for COD.
Extensive experiments on multi-modal COD tasks, integrating RGB with depth, thermal, and polarization modalities, validate the feasibility and effectiveness of our approach. As illustrated in Figure \ref{fig:Fbubble_params}, our method achieves state-of-the-art segmentation performance with the minimum number of trainable parameters compared to existing SAM-based fine-tuning approaches.
}

In summary, our contributions are as follows: 

$\bullet$ \jq{To the best of our knowledge, we are the first to present a unified framework for multi-modal camouflaged object detection that seamlessly integrates RGB with arbitrary visual modalities, including depth, thermal, and polarization.}

$\bullet$  \jq{We propose a dual-domain learning paradigm consisting of a content domain and a prompt domain. By establishing a bidirectional synergy through symmetric cross-attention, our method effectively integrates data-driven evidence and knowledge-driven priors.}

$\bullet$  \jq{Our approach demonstrates superior efficiency and scalability across diverse multi-modal segmentation datasets.}
    
\section{Related Work}

\subsection{Camouflaged Object Detection}
With the rapid development of deep learning, numerous methods have been proposed to tackle this problem, including approaches inspired by the human visual system \cite{mei2021camouflaged}, boundary-aware \cite{sun2022bgnet} models that explicitly leverage edge supervision, architectures \cite{hu2023highresolutioniterativefeedbacknetwork} that progressively refine features, frequency-domain perspectives \cite{sun2024frequency}, and diffusion-based models~\cite{10855518}.
But existing approaches still struggle in complex scenes where camouflaged objects are deeply integrated into the background, often resulting in ambiguous or missing object boundaries. Recently, to address the inherent limitations of RGB information in challenging scenarios, a growing body of research has explored multimodal approaches that integrate auxiliary modalities, including depth \cite{wang2024depth}, infrared \cite{11159525}, and polarization \cite{wang2024ipnet}. 
However, existing multimodal COD methods are typically designed for specific modality combinations and rely on customized fusion architectures. Therefore, we introduce a unified and modality-agnostic framework for camouflaged object detection, which can flexibly incorporate RGB data with diverse auxiliary modalities.

\def\w{1.0\linewidth}
\def\h{1.0in}
\begin{figure*}[tbp]
    \setlength{\tabcolsep}{1.0pt}
    \centering
    \small
    \begin{tabular}{c}
        \includegraphics[width=\w]{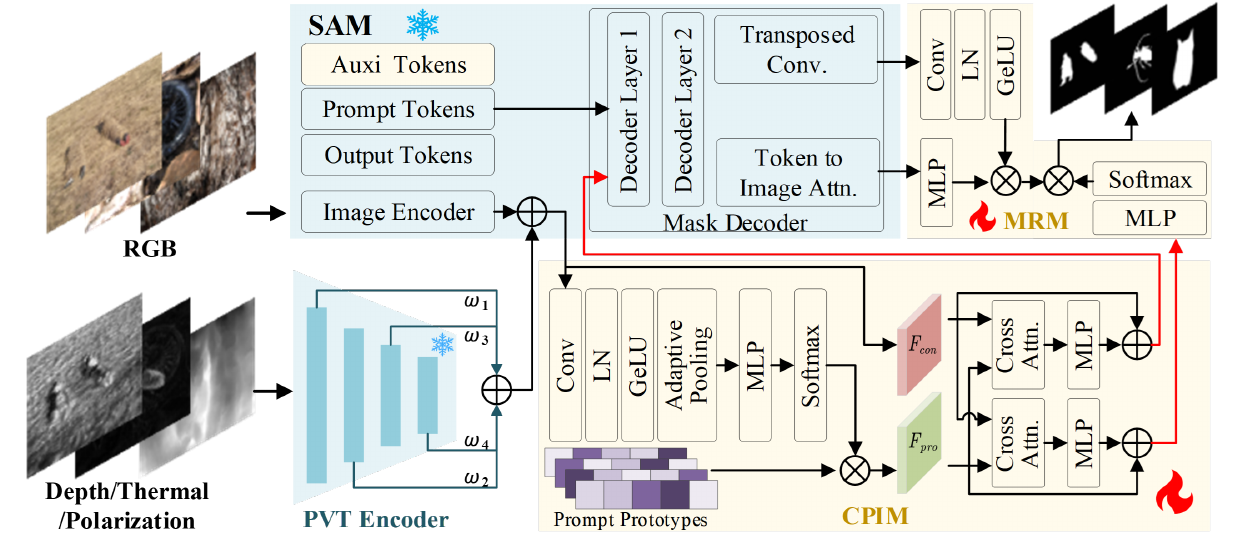}
    \end{tabular}
    \caption{Overview of our proposed framework. The framework adopts a dual-stream architecture where RGB and auxiliary modalities are processed via a frozen SAM image encoder and a PVT encoder, respectively, to construct the content domain through feature summation. In Content-Prompt Interaction Module (CPIM), a set of learnable parameters serves as the prompt domain prototype, interacting with the content domain to achieve synergy between data-driven and knowledge-driven representations. Finally, the enhanced content features, along with learnable auxi-tokens and SAM’s original tokens, are fed into the mask decoder, followed by a Mask Refine Module (MRM) to produce the final precise segmentation.}
    \label{fig:Framework}
\end{figure*}

\subsection{Segment Anything Model for COD}

SAM demonstrates strong zero-shot generalization in segmentation tasks, yet still struggles in challenging scenarios such as camouflaged object detection, motivating extensive research efforts.
One line of work focuses on adapter-based fine-tuning strategies. For example, the SAM-Adapter~\cite{chen2023sam} series of works
attributes SAM’s limitations to insufficient feature extraction in the encoder and introduces lightweight adapter modules into the image encoder. 
Another line of research explores prompt-based adaptation. For instance, COMPrompter~\cite{zhang2025comprompter} proposes a multi-prompt strategy in which boundary and box prompts mutually guide each other to enhance segmentation performance. 
Recent studies have begun to explore multimodal extensions of SAM.
For example, SAM-DSA~\cite{liu2025improving} employs a dual-stream architecture to fuse RGB and depth features for camouflaged object detection.
MM-SAM~\cite{Ren_2025_ICCV} leverages BLIP~\cite{li2022blip} to generate image captions, from which textual and visual embeddings are extracted as semantic prompts to guide SAM without manual interaction.
However, applications of SAM to multi-modal COD remain scarce, which motivates us to investigate a more general fine-tuning strategy for multi-modal settings.

\subsection{Universal Segmentation}
Unified models have emerged as a promising direction, aiming to handle diverse tasks within a single model. 
FOCUS~\cite{you2025focusuniversalforegroundsegmentation} a unified framework for foreground segmentation tasks including SOD, COD, SD, DBD, and FD.
EVP~\cite{liu2023explicit} freezes a pre-trained model and learns task-specific knowledge, providing an efficient way for unified modeling, but its performance falls behind task-specific models.
With the advent of powerful foundation models such as DINO~\cite{caron2021emergingpropertiesselfsupervisedvision} , SAM~\cite{kirillov2023segment} and CLIP~\cite{radford2021learningtransferablevisualmodels}, enabled strong generalization capabilities across diverse scenarios.
Consequently, recent efforts have shifted towards developing universal segmentation models with enhanced generalization and versatility.
VNS-SAM~\cite{guo2026vnssam} extends SAM to visually non-salient scenarios through lightweight adaptations and unified dataset.
Although unified segmentation frameworks and prompt-based approaches have shown strong capabilities, they do not explicitly address the challenge of modality-agnostic representation learning in multi-modal COD.
This motivates us to develop a unified framework that can effectively handle diverse modalities within a single architecture.

\section{Methodology}

\subsection{Overview}

\jq{Figure \ref{fig:Framework} illustrates our proposed modality-agnostic and efficient fine-tuning framework for multi-modal camouflaged object detection. The architecture employs a frozen SAM image encoder and a PVT backbone to extract features from RGB and auxiliary modalities (\textit{i.e.}, depth, thermal, or polarization), which are subsequently fused via element-wise addition to construct the content domain.
The content domain guides a set of learnable prompt prototypes to generate an adaptive prompt domain. Subsequently, a bidirectional synergy is established between the data-driven content domain and the knowledge-driven prompt domain through symmetric cross-attention.
The enhanced content features, together with learnable Auxi-tokens and SAM’s original inherent tokens, are fed into the frozen mask decoder to produce an initial mask prediction. Finally, a Mask Refine Module performs predictive calibration by integrating fine-grained prompt signals to sharpen the initial coarse masks, ensuring precise boundary segmentation.
}

\subsection{Background: SAM}
\jq{Segment Anything Model (SAM) \cite{kirillov2023segment} consists of three core components:
(i) a ViT-based image encoder that projects raw images into high-dimensional patched embeddings; (ii) a prompt encoder tasked with mapping interactive queries, such as points, bounding boxes, or masks, into a unified prompt embedding space; and (iii) a mask decoder that leverages the image embeddings along with the output and prompt tokens for final mask prediction. Benefiting from large-scale parameters and extensive data pre-training, SAM exhibits strong zero-shot transferability.
}

\subsection{Content-Prompt Interaction Module}
\jq{The Content-Prompt Interaction Module (CPIM) is pivotal to our framework for achieving modality-agnostic multi-modal COD. Rather than using modality-specific architectures or intricate multi-modal fusion schemes, the CPIM prioritizes the generation and interaction of the data-driven content domain and the knowledge-driven prompt domain. 
}

\textbf{Content Domain Construction.} 
\jq{The content domain is constructed to combine RGB information with arbitrary auxiliary visual modalities, thereby encoding target-related spatial cues and semantic information. In camouflaged object detection, relying solely on the RGB modality often proves insufficient for distinguishing targets that exhibit extreme textural similarity to their environment. To mitigate this, auxiliary modalities provide indispensable complementary cues: depth maps contribute 3D geometric structures and spatial layouts to resolve structural ambiguities; thermal infrared imaging leverages temperature differentials to highlight targets via thermal contrast; and polarization imaging captures distinctive material properties and surface orientations to bypass visual mimicry. Consequently, the fusion of these complementary cues establishes the content domain as a fundamental perception of the scene, facilitating the accurate localization and decoupling of camouflaged objects from highly complex environments.
}

\jq{Recognizing that hierarchical features are vital for COD to combine high-level semantic localization with fine-grained structural details, we obtain the RGB features $F_{rgb}$ via the element-wise addition of the embeddings from the first and last layers of the frozen SAM image encoder. Similarly, a frozen four-stage Transformer network, PVT \cite{wang2021pyramidvisiontransformerversatile}, is used to extract multi-level features for the auxiliary modality. The final auxiliary feature $F_{aux}$ is obtained by performing a weighted fusion of the outputs from all stages:
\begin{align}
   & F_{aux}^1, F_{aux}^2, F_{aux}^3, F_{aux}^4= \text{PVT} \left( I_{aux}\right), \\
& \omega_i =\sigma \!\left( \mathcal{A} \!\left(\phi (F_{\mathrm{aux}}^{i})\right) \right) , \\
 &  F_{aux}=\phi \left( \sum_{n=1}^4{\left(  \omega_i \times F_{aux}^i \right)} \right),
\end{align}
where $[\cdot]$ denotes the concatenation operation; $\phi$ represents the convolutional layer; $\sigma$ denotes the softmax function; $\mathcal{A}$ is adaptive pooling.
The final content domain features $F_{con}$ are obtained by a simple element-wise addition of the RGB features $F_{rgb}$ and the auxiliary modality features $F_{aux}$.
}

\textbf{Prompt Domain Generation.} 
\jq{Unlike the content domain encoding raw perceptual facts, the prompt domain injects high-level knowledge-driven priors into the framework. By decoupling task-specific knowledge into generic prompt representations, this domain enables the model to function without structural redesign for different auxiliary modalities, achieving modality-agnostic multi-modal COD.}

\jq{To concretely implement the aforementioned prompt domain, we first initialize a set of learnable prompt prototypes $\mathcal{P}$. These prototypes function as a global knowledge base that remains independent of the specific input data, aiming to capture the generic geometric, semantic, and material regularities of camouflaged objects. To enable these prototypes to adaptively match diverse content environments, we introduce a content-guided adaptation mechanism. Specifically, the content domain features $F_{con}$ serve as contextual cues to modulate the prototypes via a lightweight attention layer:
\begin{align}
    F_{pro} = \sigma \left( \text{MLP}\left( \mathcal{A}\left( \text{CLG} \left( F_{con} \right) \right) \right) \right)  \times \mathcal{P}, 
\end{align}
where CLG denotes an operator comprising a cascaded sequence of convolutional layers, Layer Normalization, and GeLU activation.
}

\textbf{Dual-Domain Synergistic Interaction.} 
\jq{While the content domain captures bottom-up sensory evidence encompassing multi-modal textures and spatial layouts, the prompt domain encapsulates high-level task priors and heuristic guidance. Consequently, integrating information from both domains is pivotal for accurate camouflaged object segmentation. Specifically, we employ a cross-attention mechanism to establish the correlation between them:
\begin{align}
    F_{c \leftarrow p} = \sigma \left( \frac{Q_c K_p^T}{\sqrt{d}} \right) V_p, \\
    F_{p \leftarrow c} = \sigma \left( \frac{Q_p K_c^T}{\sqrt{d}} \right) V_c,
\end{align}
where $Q, K$, and $V$ denote the linear projections of the query, key, and value tensors, respectively, with subscripts $c$ and $p$ representing the content and prompt domains. Following the dual-path interaction, the enhanced features are processed through MLP layers with residual connections to yield the final content domain output $\hat{F}_{con}$ and prompt domain output $\hat{F}_{pro}$. Subsequently, the content domain features will be fed into the SAM mask decoder, alongside auxiliary tokens, original SAM prompt tokens, and output tokens, to predict the initial mask of the target. Meanwhile, the prompt domain features will be directed to the Mask Refine Module for further refinement of the prediction. Note that the introduction of auxiliary tokens is intended to further enhance the modality-agnostic flexibility of the network.
}

\subsection{Mask Refine Module}
\jq{Previous studies \cite{chen2023sam} have demonstrated that SAM exhibits limitations in low-contrast environments, particularly when dealing with transparent or camouflaged objects where boundaries are highly ambiguous. Therefore, we propose a simple Mask Refine Module that leverages the enhanced prompt domain features $\hat{F}_{pro}$ to reconstruct the internal representations of the SAM decoder. Specifically, this module modulates the Token-to-Image Attention features ($\mathcal{F}_{attn}$) and the Transposed Convolution features ($\mathcal{F}_{conv}$), which are the two key components that jointly determine the initial prediction. Formally, the refined feature representation is computed as:
\begin{align}
    F_{mask} = \sigma(\text{MLP}(\hat{F}_{pro}))  \times (\text{CLG}(\mathcal{F}_{conv}) \times \text{MLP}(\mathcal{F}_{attn})).
\end{align}
The refined feature $F_{mask}$ will be employed to generate the final mask for the camouflaged object.
}

\begin{table*}[h]
    \centering
    \footnotesize
    \scalebox{0.95}{
        \begin{tabular}{p{2.5cm}|p{1.5cm}|p{0.5cm}p{0.5cm}p{0.5cm}p{0.5cm}|p{0.5cm}p{0.5cm}p{0.5cm}p{0.5cm}|p{0.5cm}p{0.5cm}p{0.5cm}p{0.5cm}|p{0.5cm}p{0.5cm}p{0.5cm}p{0.5cm}}
            \hline
            \hline
            \multirow{2}{*}{Methods} & \multirow{2}{*}{Pub. Year} & \multicolumn{4}{c|}{CAMO} & \multicolumn{4}{c|}{COD10K} & \multicolumn{4}{c|}{NC4K} & \multicolumn{4}{c}{CHAMELEON} \\
            \cline{3-18}
            & & $S_{\alpha}\uparrow$ & $F_{\beta}^w\uparrow$ & $E_{\phi}\uparrow$ & $M\downarrow$ & $S_{\alpha} \uparrow$ & $F_{\beta}^w\uparrow$ & $E_{\phi}\uparrow$ & $M\downarrow$ & $S_{\alpha}\uparrow$ & $F_{\beta}^w\uparrow$ & $E_{\phi}\uparrow$ & $M\downarrow$ & $S_{\alpha}\uparrow$ & $F_{\beta}^w\uparrow$ & $E_{\phi}\uparrow$ & $M\downarrow$ \\
            \hline
            \multicolumn{18}{c}{\textbf{\textit{RGB-based Methods}}}\\
            \hline
            SINet \cite{fan2020camouflaged}&CVPR'20&0.751&0.606&0.771&0.100&0.771&0.551&0.806&0.051&0.808&0.723&0.883&0.058&0.869&0.740&0.891&0.044\\
            PFNet \cite{mei2021camouflaged}&CVPR'21&0.782&0.695&0.845&0.085&0.800&0.660&0.880&0.040&0.829&0.745&0.891&0.053&0.882&0.810&0.927&0.053\\
            FSPNet \cite{huang2023feature}&CVPR'23&0.856&0.799&0.899&0.050&0.851&0.735&0.895&0.026&0.879&0.816&0.915&0.035&-&-&-&-\\ 
            FGANet \cite{zhai2022exploring}&NeurIPS'23&0.803&0.748&0.871&0.068&0.821&0.678&0.895&0.031&-&-&-&-&0.902&0.840&0.947&0.030\\
            FEDER \cite{he2023camouflaged}&CVPR'23&0.802&0.738&0.867&0.071&0.822&0.751&0.900&0.032&0.847&0.789&0.907&0.044&-&-&-&-\\
            HQSAM* \cite{ke2023segment} & NeurIPS'23&0.863&0.850&0.923&0.048&0.888&0.846&0.950&0.022&0.893&0.869&0.940&0.031&0.872&0.822&0.942&0.035\\
            SAM-Adapter* \cite{chen2023sam}& ICCV'23&0.847&0.765&0.873&0.070&0.883&0.801&0.918&0.025&-&-&-&-&0.915&0.824&0.919&0.033\\
            FocusDiff \cite{zhao2024focusdiffuser}&ECCV'24&0.812&0.772&0.883&0.069&0.820&0.730&0.897&0.031&0.850&0.810&0.902&0.044&0.890&0.843&0.938&0.028\\
            ZoomNetxt \cite{ZoomNeXt}&TPAMI'24&0.869&0.831&0.918&0.049&0.888&0.820&0.935&0.019&0.894&0.855&0.935&0.030&\textbf{0.923}&0.886&\textbf{0.958}&\textbf{0.017}\\
            SAM2-Adapter* \cite{chen2024sam2}&ICCVW'24&0.855&0.810&0.909&0.051&0.899&0.850&0.950&\textbf{0.018}&-&-&-&-&0.915&0.889&0.955&0.018\\
            GenSAM* \cite{hu2024relax}&AAAI'24&0.719&0.659&0.775&0.113&0.775&0.681&0.838&0.067&-&-&-&-&0.764&0.680&0.807&0.090\\
            MedSAM* \cite{ma2024segment}&Nature'24&0.820&0.779&0.904&0.065&0.841&0.751&0.917&0.033&0.866&0.821&0.928&0.041&0.868&0.813&0.936&0.036\\
            UTNet \cite{11175541}&TMM'25&0.677&0.518&0.733&0.123&0.699&0.485&0.773&0.068&0.743&0.603&0.798&0.086&0.818&0.709&0.885&0.054\\
            PlantCamo \cite{yang2024plantcamo}&AIR'25& 0.819& 0.757& 0.875& 0.076&0.856 &0.767 & 0.911& 0.029&0.853 &0.789 & 0.899& 0.047&0.899&0.853&0.941&0.030\\
            ESCNet \cite{ye2025escnet}&ICCV'25& 0.871& 0.843&0.934 &\textbf{0.044} &0.873 &0.804 &0.939 &0.021 &0.892 &0.859 &0.941 & \textbf{0.028}&- &- &- &- \\

            COMPrompter* \cite{zhang2025comprompter}& SCIS'25&\textbf{0.882}&0.858&0.942&\textbf{0.044}&0.889&0.821&0.949&0.023&\textbf{0.907}&0.876&\textbf{0.955}&0.030&0.906&0.857&0.955&0.026\\
            MM-SAM* \cite{Ren_2025_ICCV}&ICCV'25&0.863&0.782& 0.901 & 0.059&0.896&0.808&0.907& 0.023&-&-&-&-&\textbf{0.923} &0.853&0.946 &0.027\\
            SAM-TTT* \cite{yu2025sam}& MM'25& 0.868 & 0.838 & 0.935 & 0.045 & 0.874 & 0.805 & 0.942 & 0.027 & 0.884 & 0.837 & 0.943 &0.031&-&-&-&-\\
            ST-SAM* \cite{hu2025st}&MM'25&0.844&0.779&0.890&0.058&0.837&0.713&0.874&0.030&0.874&0.807&0.911&0.037&0.876&0.804&0.926&0.032\\  
            \hline
            \multicolumn{18}{c}{\textbf{\textit{Multimodal-based Methods}}} \\
            \hline
            SPNet \cite{zhou2021specificity}&ICCV'21&0.783&0.807&0.831&0.083&0.808&0.776&0.869&0.053&0.825&0.828&0.874&0.054&-&-&-&-\\
            SPSN \cite{leespsn}&ECCV'22&0.773&0.782&0.829&0.084&0.789&0.727&0.854&0.042&0.852&0.852&0.908&0.043&-&-&-&-\\
            DaCod \cite{wang2023depth} & MM'23 &0.856&0.796&0.905&0.051&0.840&0.729 & 0.907 & 0.028 & 0.874 & 0.814 & 0.924 & 0.035 & 0.896 & 0.835 & 0.946 &0.026\\  
            PopNet \cite{wu2023source}&ICCV'23&0.808&0.744&0.859&0.077&0.851&0.757&0.910&0.028&0.861&0.808&0.910&0.042&0.910&\textbf{0.893}&-&0.022\\
            DSAM* \cite{yu2024exploring}& MM'24 &0.832&0.794& 0.913&0.061&0.846&0.760&0.921&0.033&0.871&0.826&0.932&0.040&-&-&-&-\\
            SAM-DSA* \cite{liu2025improving}&ICCV'25 &0.866&0.839&\textbf{0.946}&0.047&0.881&0.817&0.942&0.023&0.889&0.857&0.954&0.032&0.888&0.841&0.950&0.031\\     
            \hline
            Ours-J& - & 0.863 & 0.849 & 0.867 & 0.049 & 0.890 &0.846 & 0.951 & 0.023 &0.893 & 0.868 & 0.950 & 0.031 &0.866&0.815&0.932&0.042\\
            Ours& -&0.874&\textbf{0.862}&0.931&\textbf{0.044}&\textbf{0.901}&\textbf{0.863}&\textbf{0.958}&\textbf{0.018}&0.902&\textbf{0.880}&0.943&\textbf{0.028}&0.887&0.843&0.948&0.028\\
        
            \hline
        \end{tabular}
    }
    \caption{Comparison of SOTA methods on multiple RGB-Depth datasets in terms of $S_{\alpha}$, $F_{\beta}^{w}$, $E_{\phi}$, and $M$. The best results are highlighted in \textbf{bold}. * represents the SAM-based methods.}
    \label{tab:RGBDdata}
\end{table*}

\section{Experiments}
\def\w{1.0\linewidth}
\def\h{1.0in}
\begin{figure*}[h]
    \setlength{\tabcolsep}{1.0pt}
    \centering
    \small
    \begin{tabular}{c}
        \includegraphics[width=\w]{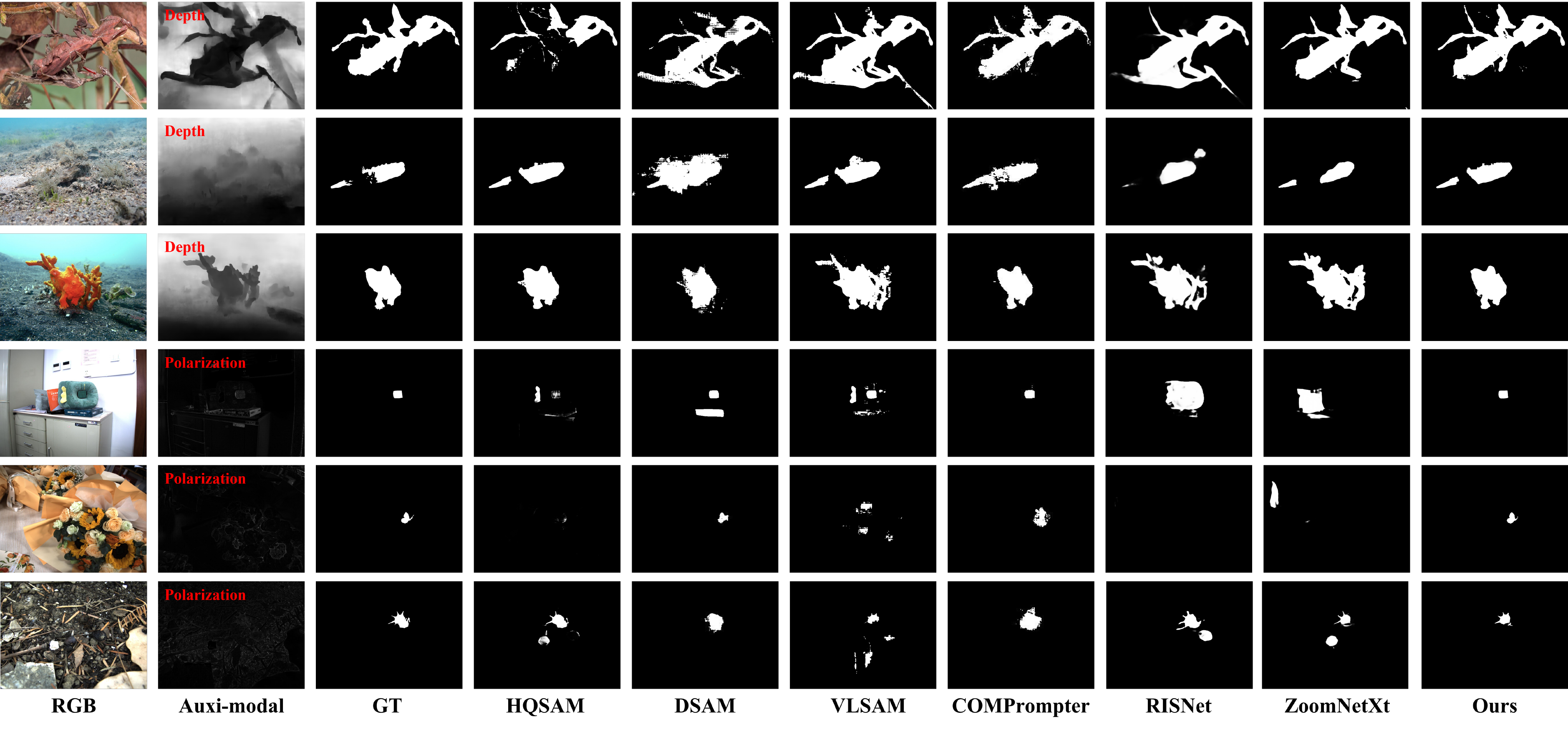}
    \end{tabular}
    \caption{Qualitative comparison of detection results on RGB-D and RGB-P camouflaged object detection datasets.}
    \label{fig:RGBD_RGBP}
\end{figure*}

\subsection{Experimental Setup}

\textbf{Implementation Details.}
Our model is implemented in PyTorch and trained on a single RTX 4090 GPU. We employ the Adam optimizer with an initial learning rate of 5e-4. Training is conducted for 50 epochs with a batch size of 4. 
\jq{Following the common practice in the SAM-based fine-tuning framework \cite{ke2023segment}, we employ the same prompt generation strategy during both training and inference, which involves generating various prompt types directly from the ground truth.}

\textbf{Datasets and Metrics.}
\jq{We conducted independent training and validation on three multi-modal COD datasets that integrate RGB with auxiliary modalities, including depth (D), thermal (T), and polarization (P).
For RGB-D COD, we adopt four widely used multimodal camouflage datasets: CAMO, COD10K, NC4K, and CHAMELEON. Following \cite{fan2020camouflaged}, the training sets of CAMO and COD10K are combined for training, while the others are used for evaluation. 
Regarding RGB-P COD, we use the PCOD-1200 dataset \cite{wang2024ipnet}, the largest real-world camouflage dataset with polarization information.
For RGB-T COD,  we employ the VIAC dataset \cite{11159525} comprising 1,500 pairs of high-quality images. We randomly selected 1,200 images from the dataset for training and 300 images for testing.
We adopt four widely recognized metrics to evaluate model performance: Mean Absolute Error ($M$), $F_\beta^w$, $S_\alpha$, and $E_\phi$.
}

\begin{table}[h]
\centering

\footnotesize
\scalebox{1.0}{
\begin{tabular}{p{2.5cm}p{1.5cm}p{0.6cm}p{0.6cm}p{0.6cm}p{0.6cm}}
\hline
Methods &Pub.'Year& $S_\alpha\uparrow$ &$E_\phi\uparrow$ & $M\downarrow$ &$F_{\beta}^w\uparrow$ \\
\hline
\multicolumn{6}{c}{\textbf{\textit{RGB-based Methods}}} \\
\hline
PFNet \cite{mei2021camouflaged}&CVPR'21 &0.876 &0.946&0.013&0.788\\
SINet-V2 \cite{fan2021concealed}&TPAMI'22& 0.882&0.941&0.013& -\\
ZoomNet \cite{pang2022zoom}&CVPR'22& 0.897&0.922&0.010& -\\
BSANet \cite{zhu2022can}&AAAI'22 & 0.903&0.945&0.011& -\\
HQSAM* \cite{ke2023segment}&NeurIPS'23&0.936&0.973&0.008&0.909\\
FPNet \cite{cong2023frequency}&MM'23 &0.880&0.945&0.013&0.799\\
FEDER \cite{he2023camouflaged}&CVPR’23&0.892 &0.946 &0.011&0.843\\
FSEL \cite{sun2024frequency}&ECCV'24 &0.934&0.978&0.006&-\\
PRNet \cite{10379651}&TCSVT'24&0.933&0.978 &0.007 &0.894\\
HGINet \cite{yao2024hierarchical}&TIP'24&0.930&0.972&0.006&-\\
ZoomNetxt \cite{ZoomNeXt}&TPAMI'24&\textbf{0.945}&0.974&0.006&0.909\\
COMPrompter* \cite{zhang2025comprompter}&SCIS'25&0.886&0.929&0.019&0.795\\
ESCNet \cite{ye2025escnet}&ICCV'25&0.925&0.969&0.007&0.881\\
UTNet \cite{11175541}&TMM'25&0.769&0.891&0.025&0.648\\
PlantCamo \cite{yang2024plantcamo}&AIR'25&0.942&0.981&0.006&0.913\\
MM-SAM* \cite{11175541}&ICCV'25&0.892&0.925&0.013&0.801\\
SAM2-UNet* \cite{xiong2026sam2}& Vis.Intell'26 &0.923 & 0.966  &0.009 &0.871\\
\hline
\multicolumn{6}{c}{\textbf{\textit{Multimodal-based Methods}}} \\
\hline
PGSNet \cite{mei2022glass} &CVPR'22 & 0.916&0.965&0.010&0.868\\
DaCOD \cite{wang2023depth}& MM'23& 0.896& 0.955 & 0.011& 0.823 \\
PopNet \cite{wu2023source}&ICCV'23 & 0.898 & 0.956 & 0.010 & 0.833\\
RISNet \cite{wang2024depth}& CVPR'24& 0.922 & 0.967 &0.008& 0.881\\
IPNet \cite{wang2024ipnet}&EAAI'24&0.922 &0.970 &0.008 &-\\
DSAM* \cite{yu2024exploring}&MM'24&0.907&0.952&0.013&0.839\\
HIPFNet \cite{wang2025polarization}&EAAI'25&0.944&0.982&\textbf{0.005}&-\\
\hline
Ours-J& - & 0.938& 0.975 & 0.008& 0.912\\
Ours & -  &\textbf{0.945}&\textbf{0.987}&\textbf{0.005}&\textbf{0.924}\\
\hline
\end{tabular}}
\caption{Quantitative comparisons with SOTAs on PCOD-1200. The best results are 
highlighted in \textbf{bold}. * represents the SAM-based methods.}
\label{tab:RGBPdata}
\end{table}

\subsection{Qualitative and Quantitative Evaluation}

\textbf{RGB-D COD.} 
\jq{As shown in Table \ref{tab:RGBDdata}, we conduct a comprehensive comparison of our method against 15 general state-of-the-art (SOTA) methods and 13 SAM-based fine-tuning approaches on RGB-D datasets. On the largest benchmark COD10K, our method achieves the best performance across all evaluation metrics, outperforming the runner-up by margins of 0.002, 0.013, and 0.008 in terms of $S_\alpha$, $F_\beta^w$, and $E_\phi$. Furthermore, our method attains the top-tier results in terms of $F_\beta^w$ and $M$ on CAMO and NC4K datasets, highlighting its effectiveness. Regarding the CHAMELEON dataset, our approach yields comparable results to existing SOTA competitors. It should be noted that, compared to other datasets, CHAMELEON is substantially less indicative due to the small set of images, containing only 76 images. Such a small sample size may constrain the model's ability to fully demonstrate its performance gains in complex scenarios. Figure \ref{fig:RGBD_RGBP} further provides a visual comparison of the results. By fully exploiting the spatial geometry and structural cues provided by the depth maps, our approach demonstrates stronger boundary adherence and structural integrity.
}

\textbf{RGB-P COD:}
\jq{On the RGB-P task, we evaluate the proposed method on the PCOD-1200 dataset against 19 general SOTA methods and 5 SAM-based fine-tuning approaches. As reported in Table \ref{tab:RGBPdata}, our method achieves state-of-the-art performance across all evaluation metrics, reaching an $S_\alpha$ of 0.945, an $E_\phi$ of 0.987, an $M$ of 0.005, and an $F_\beta^w$ of 0.924.  As illustrated by the qualitative comparisons in Figure \ref{fig:RGBD_RGBP}, polarization imaging, characterized by its high sensitivity to material properties, provides crucial complementary texture features for camouflaged targets. These quantitative and qualitative results collectively demonstrate the effectiveness and robustness of our proposed method.
}

\begin{table}[h]
\centering
\scalebox{1.}{
\begin{tabular}{p{2.5cm}p{1.5cm}p{0.6cm}p{0.6cm}p{0.6cm}p{0.6cm}}
\hline
Methods &Pub.'Year& $S_\alpha\uparrow$ &$E_\phi\uparrow$ & $M\downarrow$ &$F_{\beta}^w\uparrow$ \\
\hline
\multicolumn{6}{c}{\textbf{\textit{RGB-based Methods}}} \\
\hline
HQSAM* \cite{ke2023segment} & NeurIPS'23&0.929 & 0.981 &0.011 & 0.925\\
ZoomNetxt \cite{ZoomNeXt}&TPAMI'24&\textbf{0.934}&0.981&0.011&0.920\\
UTNet \cite{11175541}&TMM'25&0.903&0.969&0.016&0.876\\
PlantCamo \cite{yang2024plantcamo}&AIR'25&0.926&0.982&\textbf{0.010}&0.918\\
ESCNet \cite{ye2025escnet}&ICCV'25&0.928&0.979&0.011&0.919\\
COMPrompter* \cite{zhang2025comprompter}&SCIS'25&0.912&0.968&0.015&0.894\\
MM-SAM* \cite{Ren_2025_ICCV}&ICCV'25&0.916&0.972&0.013&0.906\\
SAM2-UNet* \cite{xiong2026sam2}&Vis.Intell'26 &0.929& 0.977 & 0.011 &0.915\\
\hline
\multicolumn{6}{c}{\textbf{\textit{Multimodal-based Methods}}} \\
\hline
RISNet \cite{wang2024depth}&CVPR'24&0.925 & 0.979&\textbf{0.010} &0.921\\
DSAM* \cite{yu2024exploring}& MM'24 &0.565& 0.645 &0.262&0.440\\
\hline
Ours-J& - & 0.928& 0.981 & 0.011& 0.923\\
Ours & - &0.930&\textbf{0.983}&\textbf{0.010}&\textbf{0.929}\\
\hline

\end{tabular}
}
\caption{Quantitative comparisons with SOTAs on the VIAC (RGB-T) dataset. The best
results are highlighted in \textbf{bold}. * represents the SAM-based methods.}
\label{tab:RGBTdata}
\end{table}

\begin{table*}[t]
\centering
\footnotesize
\scalebox{0.95}{
\begin{tabular}{c|l|l|cccc|cccc|cccc}
\hline
\hline
\multirow{2}{*}{Setting} & \multirow{2}{*}{\#} & \multirow{2}{*}{Method}
& \multicolumn{4}{c|}{RGB-D (COD10K)}
& \multicolumn{4}{c|}{RGB-P (PCOD-1200)}
& \multicolumn{4}{c}{RGB-T (VIAC)} \\
\cline{4-15}
& & 
& $S_\alpha\uparrow$ & $F_\beta^w\uparrow$ & $E_\phi\uparrow$ & $M\downarrow$
& $S_\alpha\uparrow$ & $F_\beta^w\uparrow$ & $E_\phi\uparrow$ & $M\downarrow$
& $S_\alpha\uparrow$ & $F_\beta^w\uparrow$ & $E_\phi\uparrow$ & $M\downarrow$ \\
\hline

\multirow{6}{*}{Standard Train}
& A & w/o WF 
& 0.897 & 0.860 & 0.958 & 0.019  
& 0.939 & 0.916 & 0.979 & 0.006 
& 0.928 & 0.926 & 0.984 & 0.010 \\

& B & w/o CPIM
& 0.887 & 0.843 & 0.950 & 0.023 
& 0.934 & 0.910 & 0.974 & 0.007 
& 0.921 & 0.917 & 0.981 & 0.012 \\

& C & CPIM w/o CA
& 0.887 & 0.845 & 0.953 & 0.021 
& 0.935 & 0.909 & 0.914 & 0.006 
& 0.919 & 0.911 & 0.979 & 0.012 \\

& D & w/o MRM
& 0.895 & 0.855 & 0.958 & 0.020  
& 0.944 & 0.923 & 0.981 & 0.006 
& 0.931 & 0.929 & 0.984 & 0.010 \\  

& E & w/o Auxi-Tokens
& 0.891 & 0.846 & 0.952 & 0.022  
& 0.943 & 0.921 & 0.982 & 0.006 
& 0.931 & 0.928 & 0.983 & 0.010 \\

& F & Ours
& \textbf{0.901} & \textbf{0.863} & \textbf{0.958} & \textbf{0.018}
& \textbf{0.945} & \textbf{0.924} & \textbf{0.987} & \textbf{0.005}
& \textbf{0.930} & \textbf{0.929} & \textbf{0.983} & \textbf{0.010} \\

\hline
\hline

\multirow{4}{*}{Joint Train}
& G & w/o Auxi-Tokens
& 0.872 & 0.815 & 0.928 & 0.030  
& 0.928 & 0.893 & 0.966 & 0.010 
& 0.915 & 0.907 & 0.975 & 0.014 \\

& H & w/o CPIM
& 0.885 & 0.835 & 0.942 & 0.025 
& 0.937 & 0.910 & 0.972 & 0.008 
& 0.916 & 0.902 & 0.975 & 0.013 \\

& I & w/o MRM
& 0.857 & 0.798 & 0.924 & 0.027  
& 0.932 & 0.911 & 0.973 & 0.010 
& 0.915 & 0.902 & 0.973 & 0.013 \\

& J & Ours
& \textbf{0.890} & \textbf{0.846} & \textbf{0.951} &\textbf{ 0.023} 
& \textbf{0.938} & \textbf{0.912} & \textbf{0.975} & \textbf{0.008 }
& \textbf{0.928} & \textbf{0.923} & \textbf{0.981} & \textbf{0.011} \\

\hline
\hline
\end{tabular}
}
\caption{Unified ablation study under standard and joint training settings.}
\label{tab:ablation_all}
\end{table*}

\def\w{1.0\linewidth}
\def\h{1.0in}
\begin{figure}[t]
    \setlength{\tabcolsep}{1.0pt}
    \centering
    \small
    \begin{tabular}{c}
        \includegraphics[width=\w]{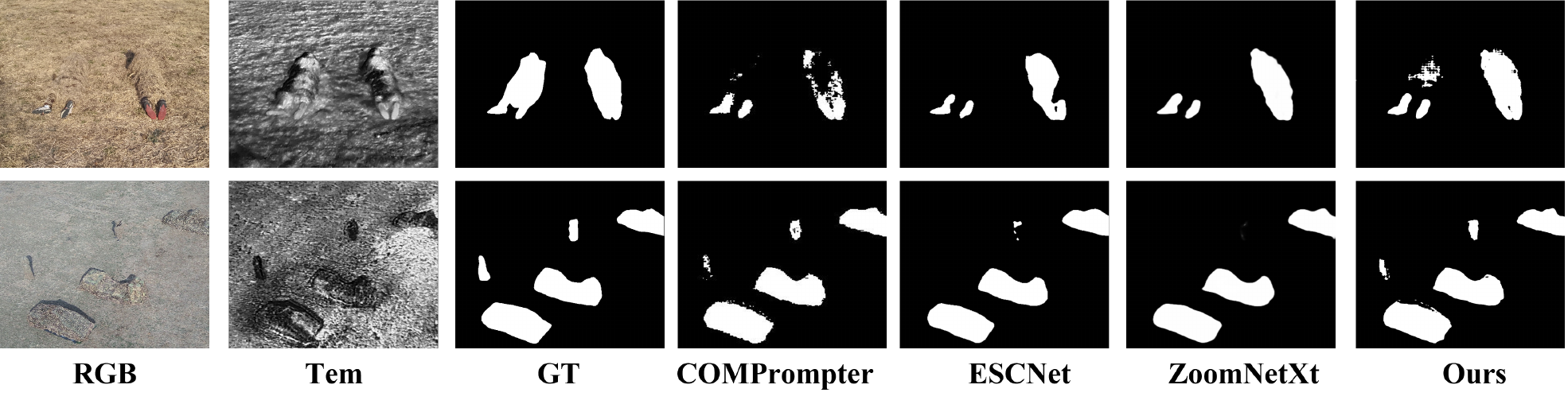}
    \end{tabular}
    \caption{Qualitative comparison of detection results on the VIAC (RGB-T) dataset.}
    \label{fig:RGBDT-COD}
\end{figure}

\textbf{RGB-T COD:}
\jq{As shown in Table \ref{tab:RGBTdata}, the evaluation on the RGB-T VIAC dataset demonstrates that our method achieves state-of-the-art performance across all metrics. Since the VIAC dataset encompasses diverse camouflage types, including plant camouflage, camouflage nets, and camouflage suits, these results thus demonstrate the exceptional generalization capability of our method across various challenging scenarios. 
Figure \ref{fig:RGBDT-COD} illustrates the qualitative segmentation results on multiple camouflaged objects. It is evident that thermal imaging, by capturing the distinct thermal signatures between targets and their surroundings, provides clear structural cues that assist our method in achieving effective and precise segmentation.
}

\textbf{Discussion.}
\jq{Prompt learning has been widely adopted as an efficient fine-tuning strategy for SAM. This work focuses on introducing prompt learning into the largely underexplored setting of multi-modal COD. Existing SAM-based methods typically rely on modality-specific designs, such as DSAM and SAM-DSA tailored for depth inputs. While effective on RGB-D benchmarks, these approaches exhibit limited cross-modal generalization. We observe that DSAM achieves competitive performance on RGB-D datasets but degrades significantly on RGB-T benchmarks. In contrast, our modality-agnostic prompting framework offers a more generalizable and scalable solution for multi-modal COD.
}

\def\w{1.0\linewidth}
\def\h{1.0in}
\begin{figure}[tbp]
    \setlength{\tabcolsep}{1.0pt}
    \centering
    \small
    \begin{tabular}{c}
        \includegraphics[width=\w]{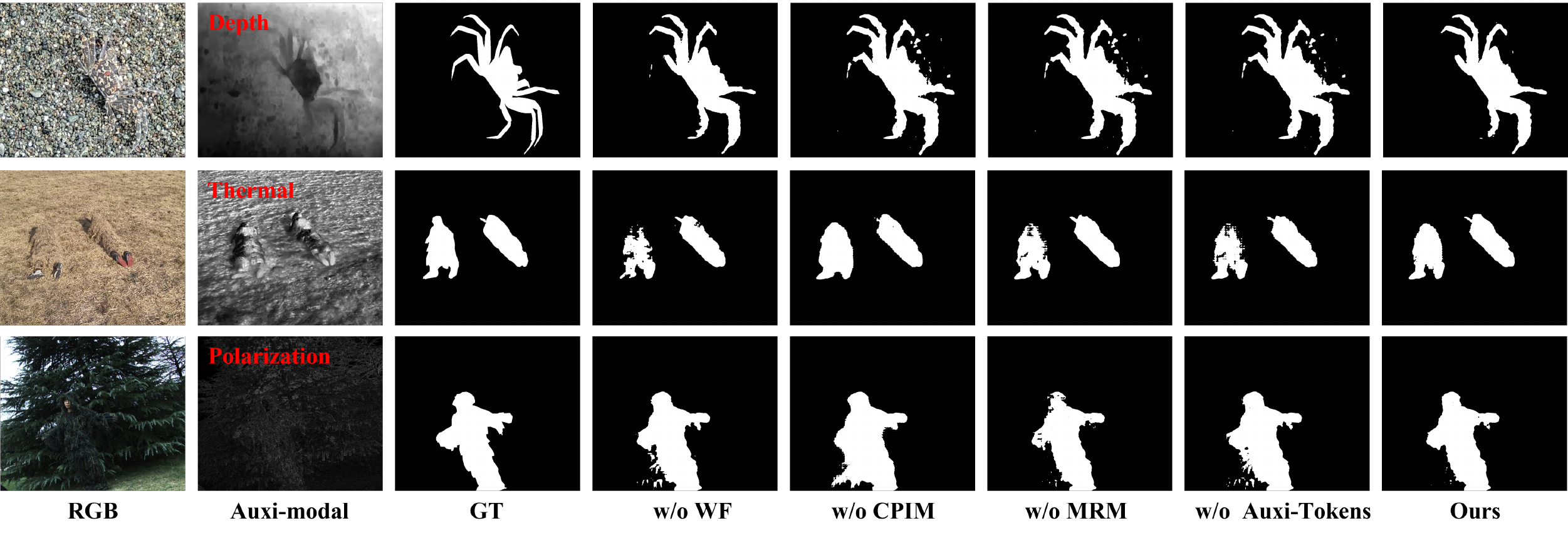}
    \end{tabular}
    \caption{Visual ablation comparison of different variants.}
    \label{fig:Ablition}
\end{figure}

\subsection{Ablation Study}

\textbf{Impact of Key Components.}
\jq{There are three components in our approach:
(i) Target content domain construction. A pivotal part of our content domain construction is the weighted fusion (WF) mechanism, which integrates multi-scale auxiliary modality features extracted by PVT. As shown in row \textit{A} of Table \ref{tab:ablation_all}, removing WF mechanism leads to a performance decline across all datasets, validating its effectiveness.
(ii) Prompt domain generation and interaction. To verify the necessity of prompt domain generation and its interaction with the content domain, we remove the entire CPIM module (\textit{i.e.}, w/o CPIM) and the Cross Attention (\textit{i.e.}, CPIM w/o CA) mechanism, respectively. The results in rows \textit{B} and \textit{C} show performance drops, indicating the vital role of CPIM in both generating the prompt domain and establishing effective correlations between content and prompt domains. 
(iii) Mask Refine Module. As shown in row \textit{D}, removing MRM leads to notable performance degradation, confirming its effectiveness in refining the initial predictions.
Moreover, row \textit{E} of Table \ref{tab:ablation_all} validates the effectiveness of the auxiliary tokens. The exclusion of these tokens leads to a performance drop across all benchmarks, suggesting that they enhance the network's flexibility and adaptability.
}

\jq{Figure \ref{fig:Ablition} further presents a qualitative comparison for the ablation study. It is evident that the MRM enhances the structural integrity of the targets. Moreover, the construction of the content and prompt domains facilitates the generation of sharper and clearer boundaries for camouflaged objects, resulting in masks that are highly consistent with the ground truth.
Figure \ref{fig:HeatMap} illustrates the heatmap visualizations under different ablation settings, demonstrating that the absence of key components results in significantly dispersed attention and a failure to precisely highlight the target regions.
}

\def\w{1.0\linewidth}
\def\h{1.0in}
\begin{figure}[t]
    \setlength{\tabcolsep}{1.0pt}
    \centering
    \small
    \begin{tabular}{c}
        \includegraphics[width=\w]{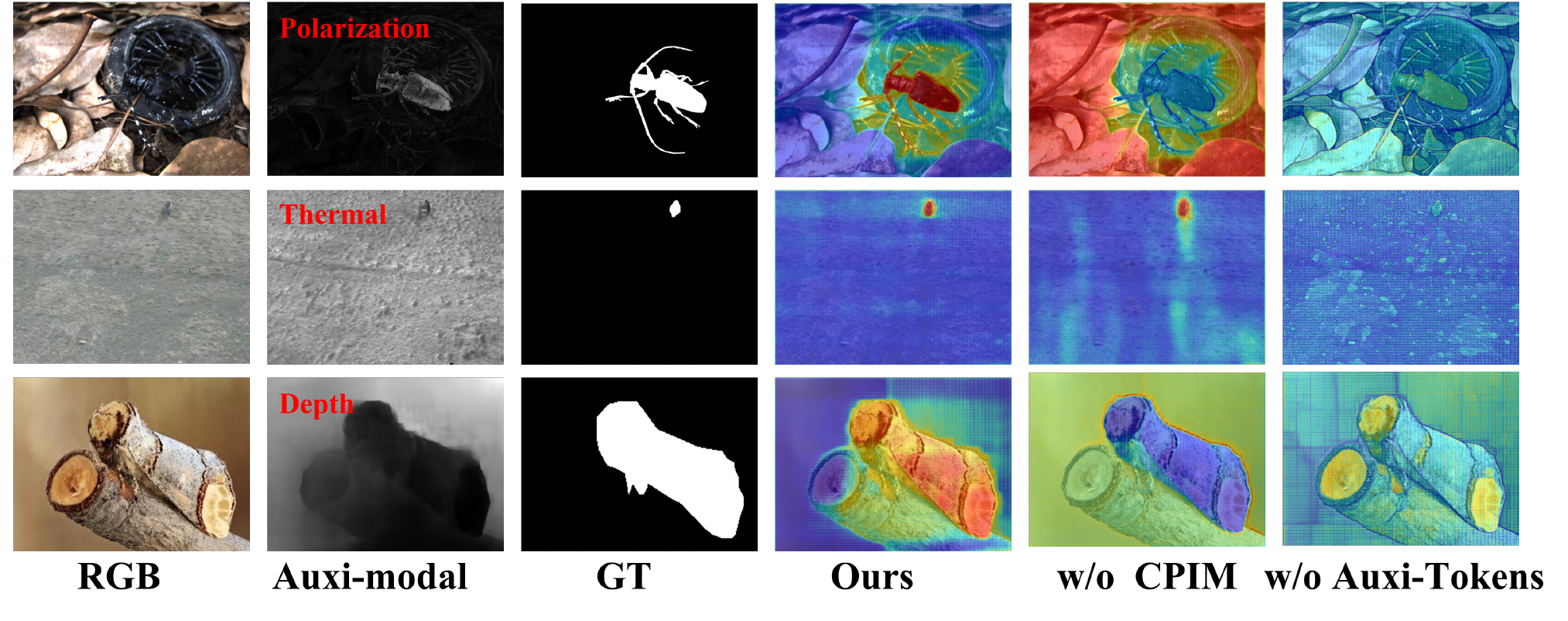}
    \end{tabular}
    \caption{Attentional visualization of the transposed convolution branch within the SAM mask decoder.}
    \label{fig:HeatMap}
\end{figure}

\begin{table}[tbp]
\centering

\setlength{\tabcolsep}{6pt}
\scalebox{1}{
\begin{tabular}{l|c|ccc}
\hline
\multirow{2}{*}{Method} 
& \multirow{2}{*}{Params (M)$\downarrow$} 
& \multicolumn{3}{c}{$F_{\beta}^{w}\uparrow$} \\
\cline{3-5}
& & COD10K & PCOD-1200 & VIAC \\
\hline
HQSAM        & 5.10   & 0.846 & 0.909 & 0.925 \\
MM-SAM       & 93.73  & 0.808 & 0.801 & 0.906 \\
COMPrompter  & 94.85  & 0.821 & 0.795 & 0.894 \\
DSAM         & 121.93 & 0.760 & 0.839 & 0.440 \\
SAM2-UNet    & 4.38   & 0.798 & 0.871 & 0.915 \\
\hline
\textbf{Ours}& \textbf{2.73} & \textbf{0.863} & \textbf{0.924} & \textbf{0.929} \\
\hline
\end{tabular}
}
\caption{Comparison of trainable parameters and performance across three multi-modal COD benchmarks.}
\label{tab:params_perf}
\end{table}

\textbf{Cross-Modality Transfer.}
To further validate the modality-agnostic property, we conduct cross-modality transfer experiments, where the model is trained with RGB and one or multiple auxiliary modalities and evaluated on unseen modalities. As shown in Table \ref{tab:cross-transfer}, performance shows only minor degradation (parentheses denote matched-modality settings) while remaining competitive. For instance, training on RGB+D and testing on RGB+P and RGB+T results in only 0.021 and 0.014 drops in $S_{\alpha}$. These results indicate that our method is not tightly coupled to any specific modality and demonstrates strong cross-modality generalization.


\textbf{Model Complexity Analysis.} 
\jq{Mainstream approaches typically fine-tune SAM to leverage its foundational segmentation capabilities, making the volume of trainable parameters a critical metric for evaluating model efficiency. As shown in Table \ref{tab:params_perf}, our method achieves superior performance across all datasets while maintaining the minimum parameter count, only 2.73M. This indicates the rationality and efficiency of our modality-agnostic prompting design.
}

\begin{table*}[h]
    \centering
    
    \footnotesize
    \scalebox{1}{
        \begin{tabular}{p{2.5cm}|p{0.55cm}p{0.55cm}p{0.55cm}p{0.55cm}|p{0.55cm}p{0.55cm}p{0.55cm}p{0.55cm}|p{0.55cm}p{0.55cm}p{0.55cm}p{0.55cm}}
            \hline
            \multirow{2}{*}{Methods} & \multicolumn{4}{c|}{VT5000} & \multicolumn{4}{c|}{VT1000} & \multicolumn{4}{c}{VT821}  \\
            \cline{2-13}
            &$S_{\alpha}\uparrow$ & $F_{\beta}^w\uparrow$ & $E_{\phi}\uparrow$ & $M\downarrow$ & $S_{\alpha} \uparrow$ & $F_{\beta}^w\uparrow$ & $E_{\phi}\uparrow$ & $M\downarrow$ & $S_{\alpha}\uparrow$ & $F_{\beta}^w\uparrow$ & $E_{\phi}\uparrow$ & $M\downarrow$ \\
            \hline
            HRTTransNet \cite{tang2022hrtransnet} & \textbf{0.945} &0.912 & 0.870 & 0.025 & 0.945 & 0.938 & 0.913 & 0.017 & 0.929 & 0.906 & 0.849 & 0.026 \\ 
            WaveNet \cite{zhou2023wavenet} &0.940&0.911&0.864&0.026&0.952&0.945&0.921&0.015&0.929&0.912&0.863&0.024\\
            SACNet \cite{wang2024alignment} & 0.933 & 0.892 & 0.838 & 0.030 & 0.949 & 0.932 & 0.907 & 0.018 & 0.917 & 0.883 & 0.817 & 0.033\\
            VSCode-v2 \cite{luo2024vscode}&0.931&-&0.962&-&0.952&-&\textbf{0.984}&-&0.932&-&0.958&-\\
            UniSOD \cite{wang2025unified}& 0.919 &0.891 &0.955 &0.021 &0.946 &0.931 &0.956 &0.013 &0.916 &0.876 &0.937 &0.024 \\
            PCNet \cite{wang2025alignment}&0.920&-&0.958&-&0.943&-&0.958&-&0.915&-&0.941&-\\
            \hline
            Ours &0.942&\textbf{0.933}&\textbf{0.978}&\textbf{0.013}&\textbf{0.954}&\textbf{0.951}&0.983&\textbf{0.011}&\textbf{0.944}&\textbf{0.934}&\textbf{0.977}&\textbf{0.012}\\
            \hline
        \end{tabular}
    }
    \caption{Comparison of SOTA methods on multiple RGB-T SOD datasets. The best results are highlighted in \textbf{bold}.}
    \label{tab:RGBT-sod}
\end{table*}

\def\w{1.0\linewidth}
\def\h{1.0in}
\begin{figure}[h]
    \setlength{\tabcolsep}{1.0pt}
    \centering
    \small
    \begin{tabular}{c}
        \includegraphics[width=\w]{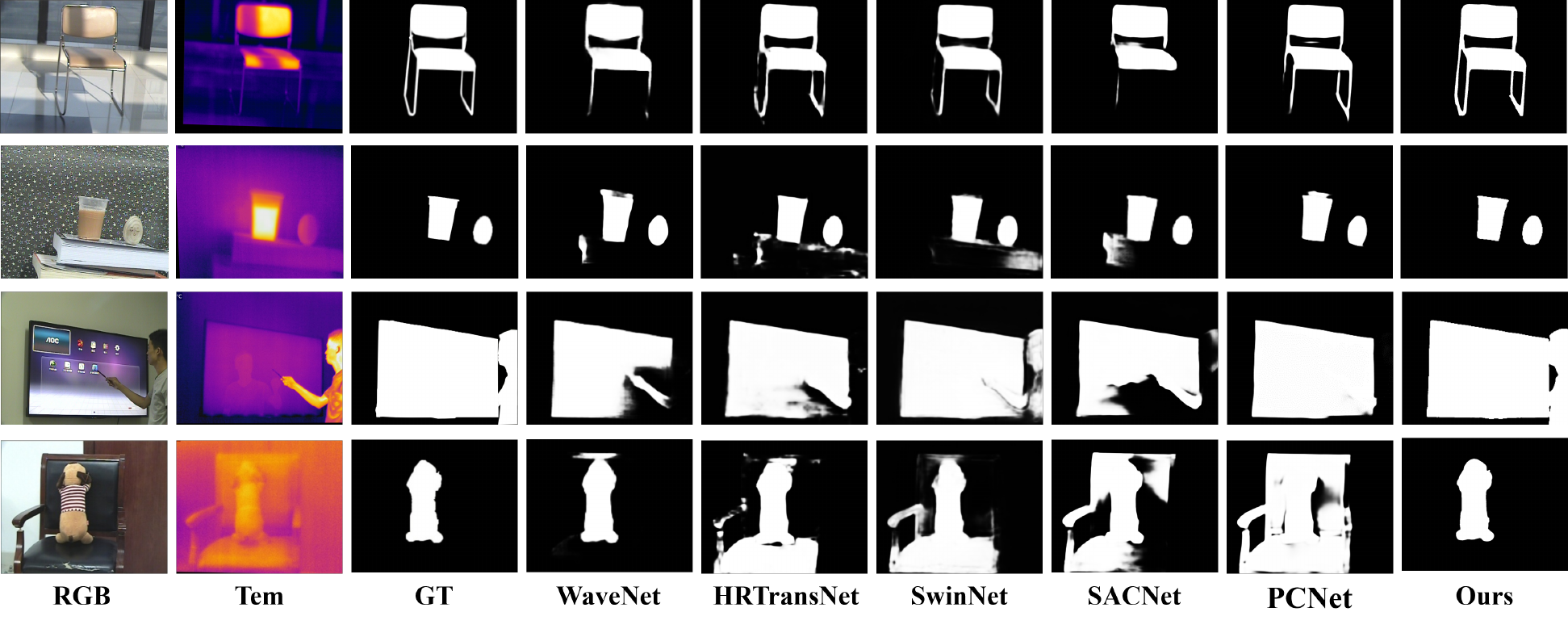}
    \end{tabular}
    \caption{Qualitative comparison of detection results on RGB-T SOD datasets.}
    \label{fig:RGBDT-SOD}
\end{figure}

\begin{table*}[h]
\centering
\small
\begin{tabular}{c|c|cccc}
\hline
\textbf{Train Modality} & \textbf{Test Modality} &$S_\alpha\uparrow$ &$E_\phi\uparrow$ &$F_{\beta}^w\uparrow$  &$M\downarrow$ \\
\hline
RGB+D   & RGB+P   & 0.924 (0.945) & 0.968 (0.987) & 0.890 (0.924) & 0.010 (0.005)\\
RGB+D   & RGB+T   & 0.916 (0.930) & 0.976 (0.983) & 0.908 (0.929) & 0.013 (0.010)\\
RGB+DT  & RGB+P   & 0.923 (0.945)& 0.975 (0.987) & 0.889 (0.924) & 0.009 (0.005) \\
RGB+DP  & RGB+T   & 0.920 (0.930) & 0.978 (0.983) & 0.915 (0.929) & 0.012 (0.010)\\
\hline
\end{tabular}
\caption{Quantitative comparison of cross-modality transfer experiments (parentheses denote matched-modality settings).}
\label{tab:cross-transfer}
\end{table*}

\textbf{Impact of Joint Training.} 
\jq{A natural conjecture is whether a unified training strategy, which aggregates datasets from all modalities, could further enhance network performance by leveraging increased data diversity. To explore this, we conducted a series of joint training experiments. As indicated by \textbf{Ours-J} in Tables \ref{tab:RGBDdata}, \ref{tab:RGBPdata}, and \ref{tab:RGBTdata}, the joint training results exhibit a slight performance decline across all benchmarks, yet still yield comparable results. We attribute this primarily to the significant scale imbalance between datasets, where the training set for RGB-D exceeds 4,000 images while the RGB-P dataset contains only 970 samples.
}

We further conduct ablations under the joint training setting. The results in Table \ref{tab:ablation_all} show consistent conclusions with those obtained under separate training, indicating that the effectiveness of each component is stable across different training paradigms.

\def\w{0.95\linewidth}
\def\h{1.0in}
\begin{figure}[tbp]
    \setlength{\tabcolsep}{1.0pt}
    \centering
    \small
    \begin{tabular}{c}
        \includegraphics[width=\w]{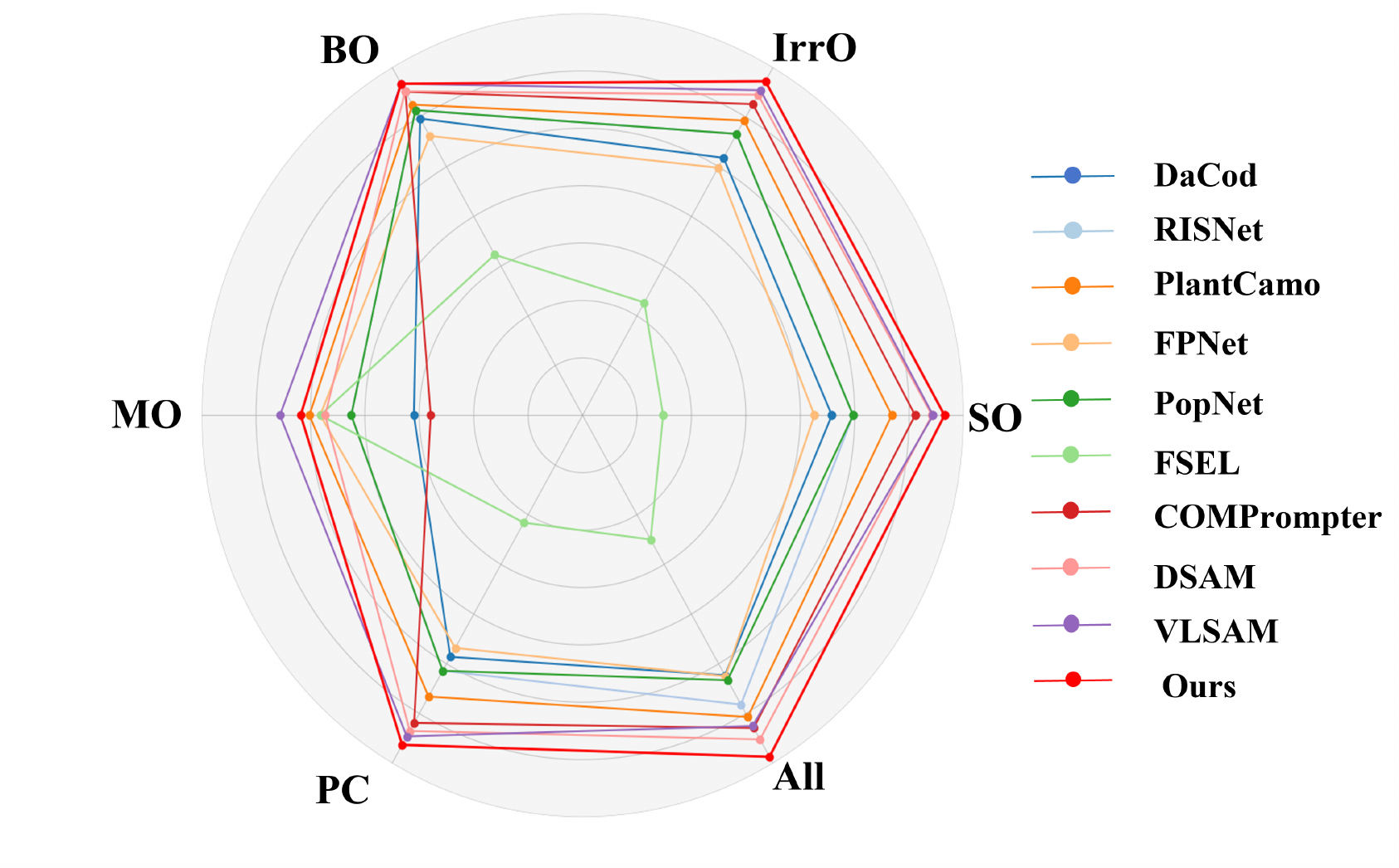}
    \end{tabular}
    \caption{Comparison under different challenging scenarios.}
    \label{fig:rader}
\end{figure}

\textbf{Performance on Different Attributes.} 
\jq{  
Figure \ref{fig:rader} compares performance under different scenarios on the PCOD-1200 dataset, including SO (Small Objects), IrrO (Irregular Objects), MO (Multiple Objects), BO (Big Objects), and PC (Protective Coloration). 
Our method achieves competitive results across challenging scenarios, benefiting from its ability to fuse information from RGB and auxiliary modalities.
} 



\subsection{Generalizability Analysis}
\jq{To further evaluate the versatility of our proposed method beyond the multi-modal COD task, we extend our evaluation to multi-modal Salient Object Detection (SOD). As shown in Table \ref{tab:RGBT-sod}, our method significantly outperforms competing methods on most standard evaluation metrics across multiple RGB-T SOD benchmarks.
This performance gain demonstrates the strong cross-domain generalization ability of our framework when transferred from COD to SOD tasks. 
The qualitative results presented in Figure \ref{fig:RGBDT-SOD} further indicate the effectiveness of our segmentation approach.
}

\def\w{1.0\linewidth}
\def\h{1.0in}
\begin{figure}[htbp]
    \setlength{\tabcolsep}{1.0pt}
    \centering
    \small
    \begin{tabular}{c}
        \includegraphics[width=\w]{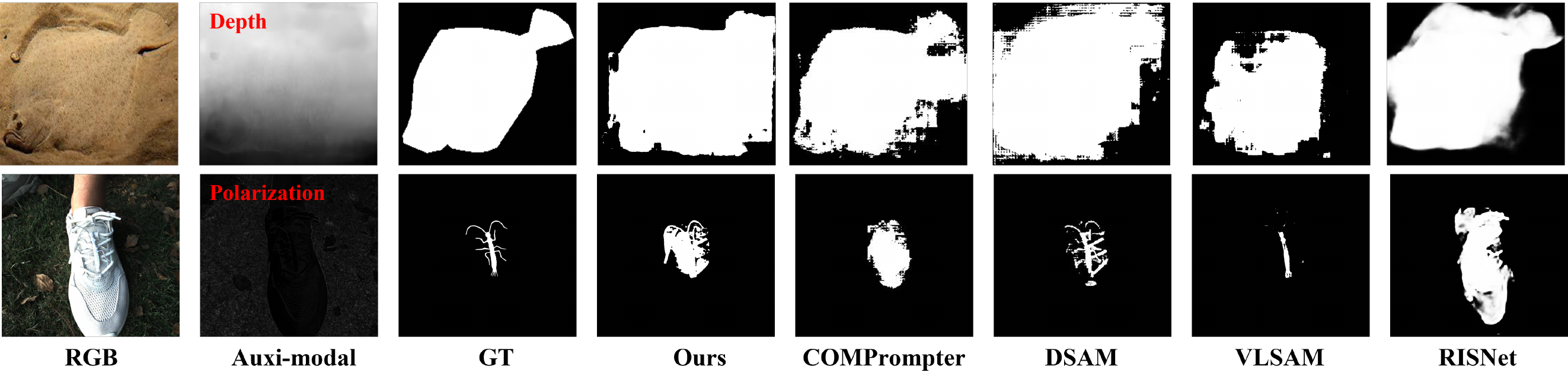}
    \end{tabular}
    \caption{The case of failure.}
    \label{fig:Limit}
\end{figure}

\subsection{Limitations}
\jq{The performance of our approach may be constrained when auxiliary modalities fail to yield informative cues or introduce noise. As shown in Figure \ref{fig:Limit}, in scenarios where RGB images lack discriminative textures and auxiliary modalities provide insufficient complementary data, it becomes difficult for all methods to detect camouflaged targets.
}

\section{Conclusion}
This paper proposes a modality-agnostic multimodal prompt framework for camouflaged object detection based on the Segment Anything Model. 
Our approach adopts a dual-domain learning paradigm consisting of a content domain and a prompt domain, which jointly capture data-driven perceptual evidence and knowledge-driven priors to achieve effective and flexible multimodal camouflaged object detection. Besides, a mask refine module is further employed to improve boundary accuracy.
Extensive experiments on diverse modality combinations validate the effectiveness of the proposed approach.
Future work will explore a unified multi-modal foreground segmentation framework that covers COD, SOD, shadow separation, and beyond.
\bibliographystyle{IEEEtran}
\bibliography{reference}{}

\end{document}